# An Amphibious Fully-Soft Miniature Crawling Robot Powered by Electrohydraulic Fluid Kinetic Energy

*Quan Xiong, Xuanyi Zhou, Jonathan William Ambrose, Raye Chen-Hua Yeow\**

Q. Xiong, X. Zhou, J. W. Ambrose, R. C.-H. Yeow
Department of Biomedical Engineering
National University of Singapore
15 Kent Ridge Cres, 119276, Singapore
E-mail: rayeow@nus.edu.sg

Miniature locomotion robots with the ability to navigate confined environments show great promise for a wide range of tasks, including search and rescue operations. Soft miniature locomotion robots, as a burgeoning field, have attracted significant research interest due to their exceptional terrain adaptability and safety features. In this paper, we introduce a fully-soft miniature crawling robot directly powered by fluid kinetic energy generated by an electrohydraulic actuator. Through optimization of the operating voltage and design parameters, the crawling velocity of the robot is dramatically enhanced, reaching 16 mm/s. The optimized robot weighs 6.3 g and measures 5 cm in length, 5 cm in width, and 6 mm in height. By combining two robots in parallel, the robot can achieve a turning rate of approximately 3 degrees/s. Additionally, by reconfiguring the distribution of electrodes in the electrohydraulic actuator, the robot can achieve 2 degrees-of-freedom translational motion, improving its maneuverability in narrow spaces. Finally, we demonstrate the use of a soft water-proof skin for underwater locomotion and actuation. In comparison with other soft miniature crawling robots, our robot with full softness can achieve relatively high crawling velocity as well as increased robustness and recovery.

## 1. Introduction

Miniature locomotion robots operating at the micro- to centimeter length scales demonstrate remarkable locomotion and navigation capabilities within highly confined and unstructured spaces [1–4]. Their unique abilities have unlocked tremendous potential in various tasks, such as drug delivery, collection, search and rescue operations [5,6]. Over the past few years, there has been a significant surge in research interest surrounding soft locomotion robots, primarily due to their exceptional terrain adaptability and inherent safety features [7–11]. These robots are built with components made from deformable and compliant materials [12–14], allowing them to accommodate diverse surroundings and interact safely with uncertain environments.

Actuators play a critical role in determining the softness and performance of miniature robots. Generally, high bandwidth and power-weight ratio are essential requirements for actuators in miniature locomotion robots. While some traditional rigid actuators, such as vibrating motors and piezoelectric actuators, easily meet these requirements [15–17], conventional soft actuators face significant challenges. Pneumatic actuators, as the predominant choice for soft actuators, have been widely utilized in soft miniature locomotion robots due to their simple structure and fabrication process [18–20]. Nevertheless, their bandwidth is limited [21,22] (typically less than 5 Hz), and the dimension and weight of air tubes impede their miniaturization and hinder long-distance locomotion. In response to these limitations, alternative soft actuators, such as dielectric elastomer actuators (DEAs) and shape memory alloys (SMAs), have been proposed. DEA-based soft miniature robots exhibit high maneuverability [23–25]. However, their fabrication is complex due to the pre-stretching frame mechanism and the challenges posed by the miniature scale, and the risk of an electric breakdown may result in severe damage to the robot [26,27]. On the other hand, SMA actuators have a slow dynamic response due to the long cooling and recovery time required by the alloy coil to return to its initial state, thereby limiting the robot's overall performance [28,29].

The emerging soft electrohydraulic actuator, known as the hydraulically amplified self-healing electrostatic (HASEL) actuator [30], exhibits exceptional characteristics, including rapid response and a high power-weight ratio[31]. Electrohydraulic actuators harness the Maxwell stress[32] generated by electrostatic fields to compress the dielectric fluid within a bladder, utilizing the resulting deformation of the bladder as a source of mechanical actuation. This bladder deformation mechanism of electrohydraulic actuators has found widespread application in various robotic systems, including robotic grippers [33–35], locomotion robots [36,37], and artificial muscles[31,38]. However, the investigation and application on the directly-generated kinetic energy of the dielectric fluid, which arises from its compression by the Maxwell stress, remains relatively limited and unexplored.

In this study, we present a fully-soft miniature crawling robot (MCR) directly powered by electrohydraulic fluid kinetic energy (EFKE). The MCR here consists of only a single soft body that is lightweight and compact. We use an electrohydraulic actuator to generate unidirectional

EFKE, such that its liquid inertia propels the MCR along the ground. To extend the robot's mobility to underwater environment, we have designed a soft water-proof skin that encapsulates the HASEL actuator, capitalizing on this novel EFKE-driven mechanism for aquatic locomotion. Furthermore, we introduce reconfigurability into the MCR, enhancing its maneuverability, and basically propose two simple methods for configuring the robot to achieve locomotion with multiple degrees of freedom (DOF). To improve convenience and reliability, we utilize a commercially available pre-coated biaxially-oriented polypropylene (BOPP) membrane in the fabrication of the HASEL actuator, effectively preventing issues like high melting temperature and edge electric breakdown. Additionally, we have optimized the MCR's design parameters and operating voltage to achieve higher motion velocity. Notably, the fully-soft architecture of the MCR has remarkable robustness and recovery, enabling the robot to continue functioning after enduring significant mechanical impacts.

## 2. Results and Discussion
### 2.1. Principle and Design Rationale

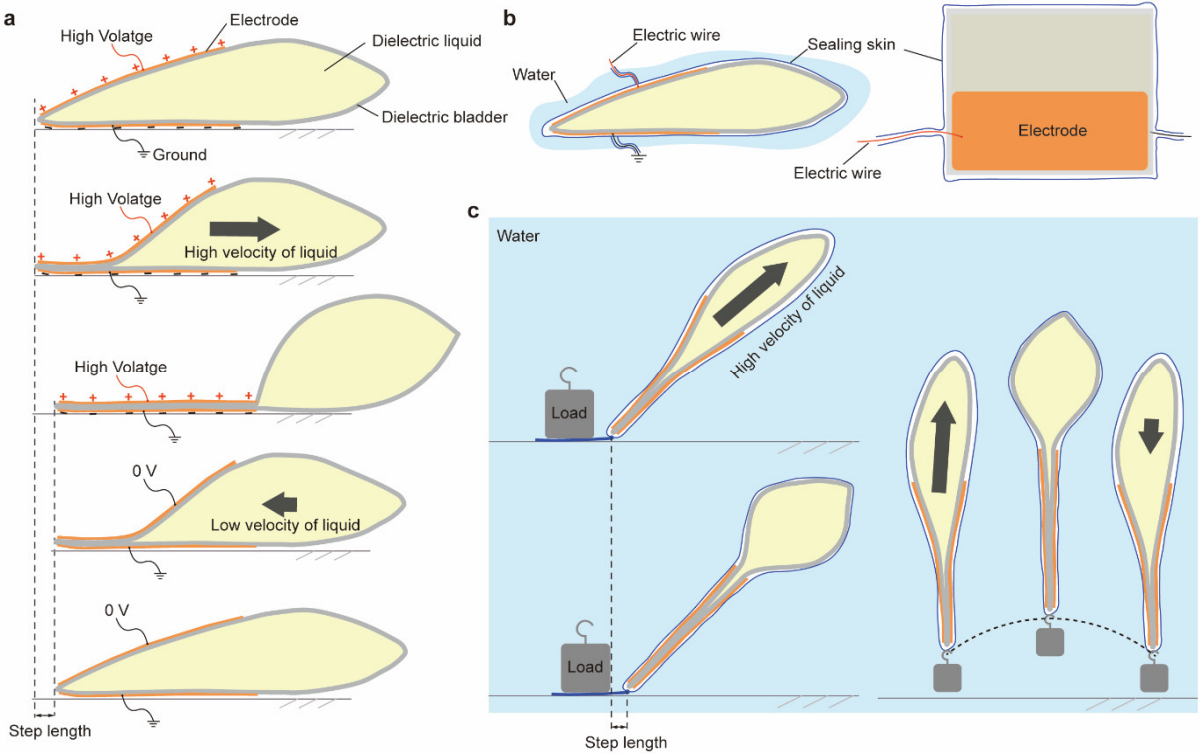

**Figure 1.** The principle and design of the MCR. (a) Principle of the MCR. (b) The water-proof skin. (c) Our MCR with the skin can achieve crawling and vertical jumping underwater.

The electrohydraulic actuator here is composed of two flexible electrodes, a deformable dielectric bladder and dielectric liquid enclosed within the bladder, as shown in **Figure 1**a. The soft dielectric bladder and liquid are sandwiched between two electrodes. Supplied with high voltage (HV), positive and negative charges are induced in the two electrodes respectively, which acts as a charging capacitance. Due to the Maxwell stress generated by the HV, the two electrodes compress the bladder and transfer kinetic energy (EFKE) to the internal liquid. The Maxwell stress $\sigma_M$ is related to the dielectric permittivity $\varepsilon$ and the electric field intensity $E$:

$$\sigma_M = \frac{1}{2}\varepsilon E^2. \tag{1}$$

To generate unidirectional EFKE, two electrodes are deployed on one side, so that the EFKE can push the MCR towards the other side (Movie S1). According to Equation 1, the Maxwell stress is highest at the edge of our MCR, so the bladder is zipped by the electrodes (Figure 1a). The bladder is made from inextensible membranes without bending stiffness, hence it is unable to absorb energy. Most of the EFKE is consumed by the dynamic friction between the ground and the MCR, and the residual EFKE is converted to the gravitational potential energy of the liquid. Removing the high electrostatic fields, the dielectric liquid flows slowly to its initial position because of its gravitational potential energy (Figure 1a). The reflux liquid with low kinetic energy cannot propel the MCR. By repeating these processes, our MCR can crawl continuously.

Miniature locomotion robots capable of amphibious mobility have a great advantage in unpredictable field environments and weather. More importantly, a humid or water environment may bring about short-circuit and electric breakdown that cause huge damage to the electrohydraulic actuator. Hence, we designed a soft water-proof skin made from the same membrane as the bladder, wrapping and fitting the MCR (Figure 1b). The water-proof skin allows our MCR to crawl in the water without affecting its mobility since the skin can deform synchronously with the bladder without energy absorption. For underwater crawling, a load is connected to the tail of the MCR (Figure 1c), The load is required to improve the underwater locomotion and to prevent the MCR from floating up since water is denser. Even if the MCR is tilted at a certain angle with the ground by its buoyancy, the EFKE is still able to propel the robot to crawl underwater (Movie S2). We also illustrated the MCR vertical jumping capability under EFKE actions by reducing the weight of the load (Movie S3).

**2.2. Fabrication based on Pre-coated BOPP Film**

To enhance convenience and reliability, we use pre-coated BOPP film to fabricate the electrohydraulic actuator (MCR). The pre-coated BOPP film is a composite of a BOPP membrane (16 μm thick) and a layer of ethyl vinyl acetate (EVA) adhesive (9 μm thick). Firstly, we cut the pre-coated BOPP membrane into 7 cm × 7 cm rectangles, which are stacked on top of each other with the EVA adhesive layers facing each other (**Figure 2**a). Secondly, an iron is used to seal the edge of the bladder by melting the EVA adhesive, but leaving a port for liquid injection. The iron temperature is about 100 °C and the wielding width of the edge is 1 cm. Thirdly, we attach two conductive tapes (copper tapes with 50 μm thickness and 5 cm width) as electrodes on the BOPP surfaces. Fourthly, we inject the dielectric oil into the bladder and seal the port. Lastly, we connect two electric wires to the electrodes. To fabricate the water-proof skin, we use two pro-coated BOPP membranes covering the MCR. By a similar path, we seal the skin along the shape of MCR and the electric wires (Figure 2b).

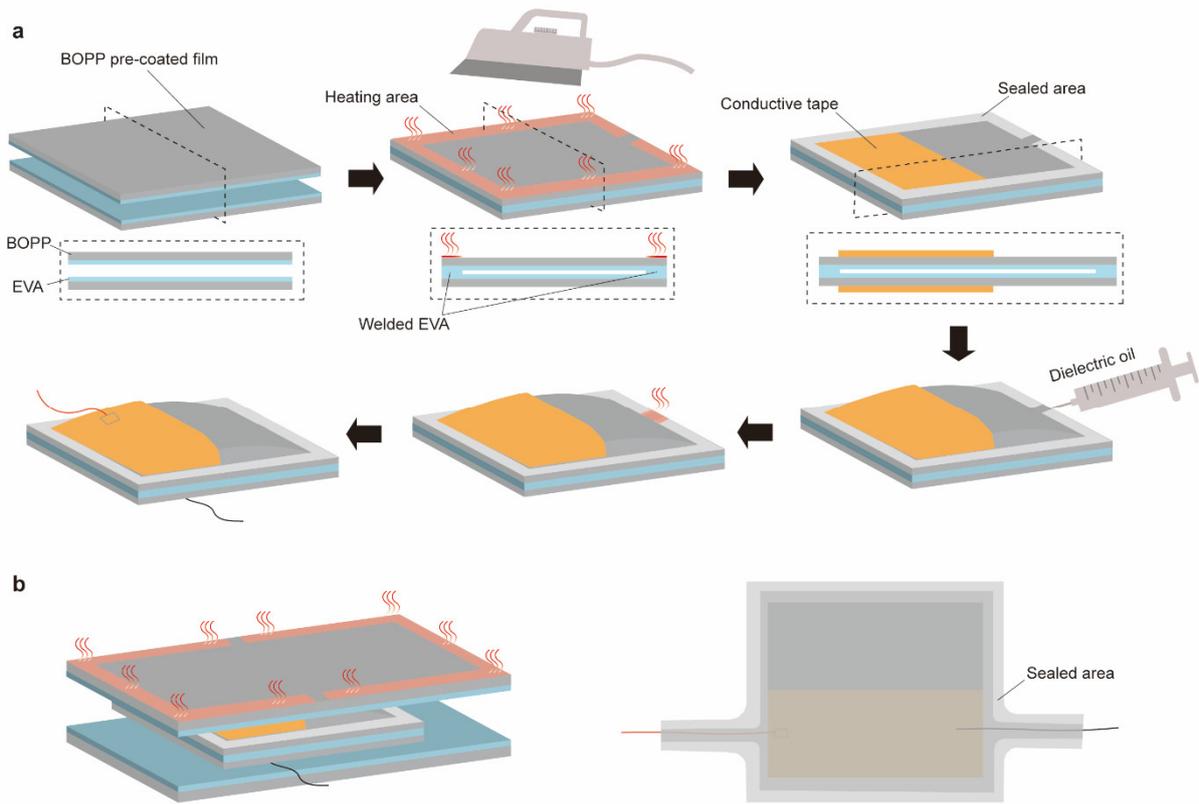

**Figure 2.** The fabrication process of the MCR. (a) The detailed fabrication of the electrohydraulic actuator. (b) The fabrication of the water-proof skin.

In contrast with previous BOPP-based fabrication methods[31], the major advantage of our approach is using lower heating temperatures to seal the bladder. Since the melting point of BOPP is approximately 170 °C [39], overhigh heating temperature will melt the BOPP layer and the liquidity of melted BOPP material may result in uneven thickness at the welding edge which increases the likelihood of electric breakdown. Furthermore, the commercialized copper tape enables us to easily fabricate electrodes and adjust their position.

### 2.3. Optimization of Crawling Velocity

The crawling velocity is dependent on the EFKE and the operating frequency according to the principle of our MCR. Although higher voltage, in theory, can produce higher EFKE propelling the MCR further, the operating frequency will be restricted by the time it takes for the dielectric liquid to return to its initial position. The whole operation period consists of zipping time (ZT, the period with HV) and releasing time (RT, the period without HV). After removing the HV, the electroadhesion between two electrodes cannot disappear immediately and need a residual time related to the amplitude and duration of HV to recover[40,41]. If we reactivate the HV before the liquid return, the EFKE decreases. This is because little liquid flows back between the two electrodes and the work done by electroadhesion drops down.

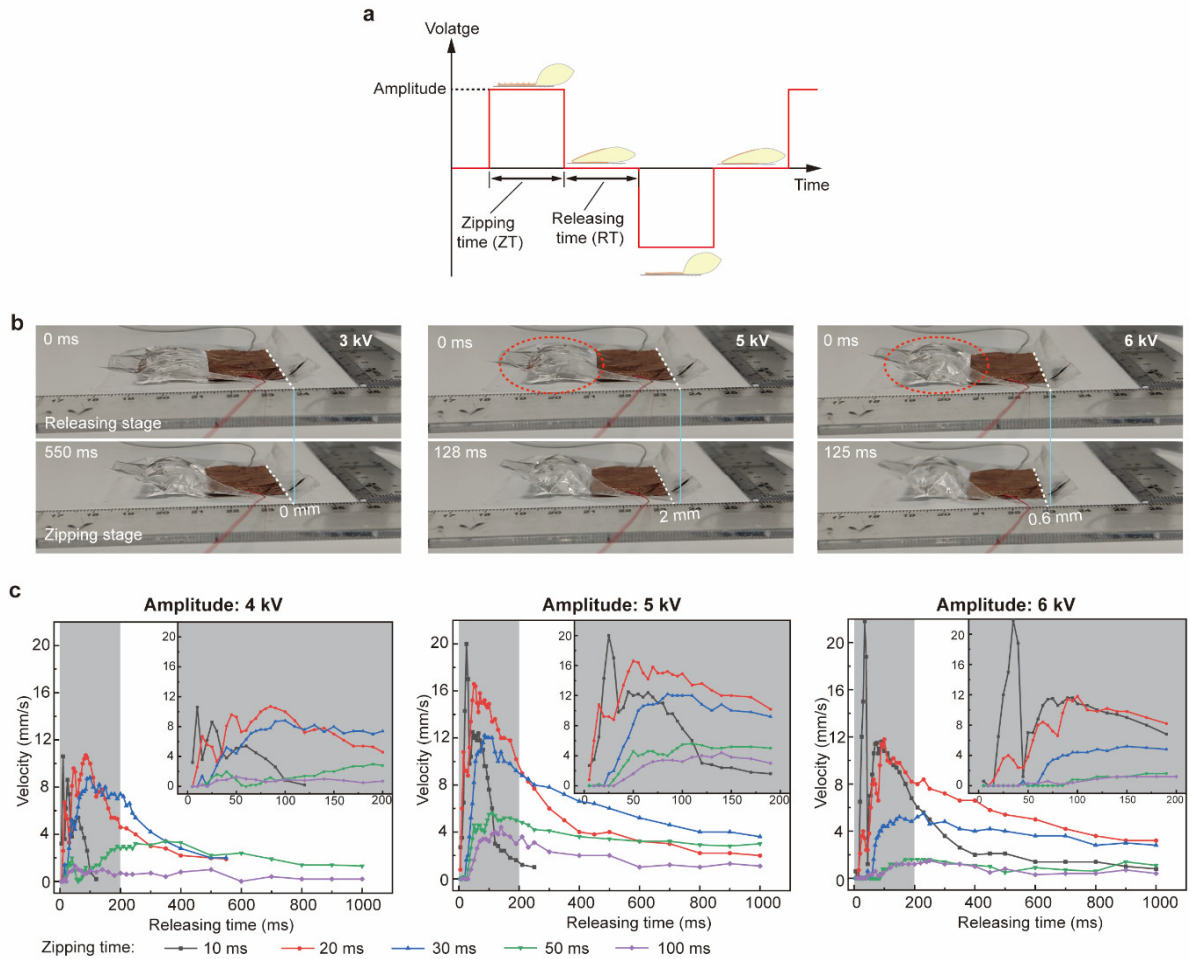

**Figure 3.** Optimization of Crawling Velocity. (a) AC operating HV. (b) The stride with different amplitudes of HV. (c) The experimental crawling velocity with various ZT/RT operating patterns and amplitudes of HV.

To reduce the residual charge, we adopt AC HV as shown in **Figure 3**a. The down electrode is always grounded, and the up electrode is connected to the HV port which can output a positive or negative HV. Firstly, we find the amplitude of the AC HV dramatically influences the stride (Movie S4). With a 3 kV HV (ZT: 2s; RT: 2s), the MCR is almost stationary, whereas it achieves a 2 mm stride with a 5 kV HV. As the HV increases to 6 kV, the stride drops to 0.8 mm because residual electroadhesion hinders the reflux as mentioned (red oval area, Figure 3b). We first have to optimize the ZT and RT that indicate the frequency and duty cycle of the HV, directly for crawling velocity. We control the amplitude of HV to 4 kV, 5 kV and 6 kV respectively, and measure the velocity of the MCR at different ZT and RT patterns. In this experiment, the ZT is set to five alternative values (10, 20, 30, 50 and 100 ms) and we gradually increase the RT from 5 to 1000 ms. The crawling velocities of all ZTs initially increase and then decrease with increasing RT. As illustrated in Figure 3c, it reaches nearly the maximum value within 200 ms of RT, so we conducted more detailed tests within this range (gray areas in Figure 3c). With 10 ms ZT, the velocity of all HV amplitudes dramatically fluctuates within 50 ms of RT, and then reaches a relatively high and stable plateau. With 20 ms ZT, the velocity increases with fluctuation initially and then reaches its maximum value which is higher than that plateau of 10 ms ZT. After that, the crawling velocity slowly decreases with RT. With 30 ms ZT, the velocity rises smoothly with RT in the beginning, followed by a gradual descent too. The velocity of 50 and 100 ms ZT is much lower than others within 200 ms of RT. Based on the specific HV amplitudes and ZTs, we can choose corresponding RT values to achieve high and stable crawling velocity. According to the above criteria, we selected optimized ZT/RT operating patterns for different amplitudes of HV: 10/60 ms, 20/80 ms, and 30/120 ms for 4 kV and 5 kV HV; 10/80 ms, 20/120 ms, and 30/180 ms for 6 kV HV.

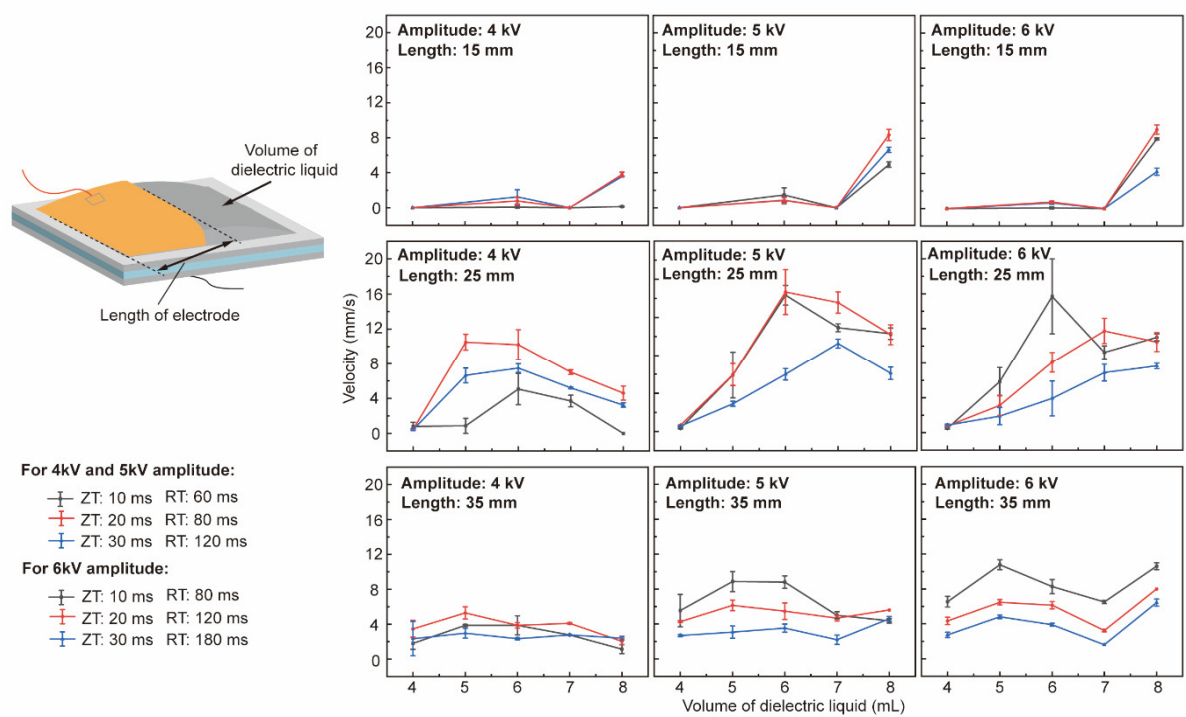

**Figure 4.** The MCR's velocity with alternative design parameters and amplitudes of HV.

Based on the selected ZT/RT patterns, we proceed to optimize the amplitude of HV, the length of the electrode and the volume of the dielectric liquid of the MCR (**Figure 4**). We fabricated MCRs with various length of electrodes at 15 mm, 25 mm, and 35 mm at 5 cm widths. In addition, various volumes (4 mL, 5 mL, 6 mL, 7 mL and 8 mL) of dielectric oil were explored.

With a 15 mm length of electrode, all crawling velocities under all HV patterns are less than 2 mm/s, if the volume is under 7 mL. When the volume surpasses 7 mL, the velocity is from 3 to 8 mm/s. With a 25 mm length of electrode, MCR's average velocity improved significantly more than others. They exhibited the highest crawling speed of 16 mm/s with 5 kV HV (ZT/RT: 20/80 and 10/60 ms) and 6 mL oil. With a 35 mm length of electrode, the velocities fall between of 15 mm length and 25 mm length data and show no obvious trends. Based on these experimental data, we selected the 25 mm long electrode, the 6 mL dielectric oil and the 5 kV amplitude of HV. The optimized MCR weighs 6.3 g, and it is 5 cm long, 5 cm wide and 6 mm high (maximum value in operation).

**2.4. Design Reconfiguration for 2-DoF Locomotion**

Based on the above-optimized design and operating HV, the MCR can crawl at a fast speed (Movie S5). Even dragging a load, it can also achieve effective locomotion (**Figure 5**a). We tested the crawling velocity of the MCR with different loads from 0 to 20 g (317% of the weight of MCR). The results (Figure 5b) demonstrated that the velocity decreases with the load from 12.7 mm/s to 0.94 mm/s.

To improve its maneuverability, we reconfigured the MCR to enable 2-DoF locomotion in a plane. One approach is to connect two MCRs in parallel. By applying the same HV in the two MCRs, the 2-DoF robot moves forward (Figure 5c). If only the left MCR is activated, the robot can turn right at 3.46 degree/s. Reversely, it turns left at 2.86 degree/s with the activated right MCR (Movie S6). Another approach is to reconfigure the distribution of electrodes on the MCR. Here, we split the MCR into four areas and deploy four corresponding pairs of electrodes (Figure 5d). The MCR can generate EFKE in 4 directions and enable the MCR to crawl multi-directionally by activating different electrodes as shown in Figure 5e and Movie S7. Compared with the first approach, the 4-electrode-MCR requires only one electrohydraulic actuator making it more compact. Therefore, it has advantages, especially in confined spaces for rescue tasks. To show its potential, we controlled the 2-DoF MCR transporting medication while passing through a winding gap (1 cm) to the injured (Figure 5f and Movie S8).

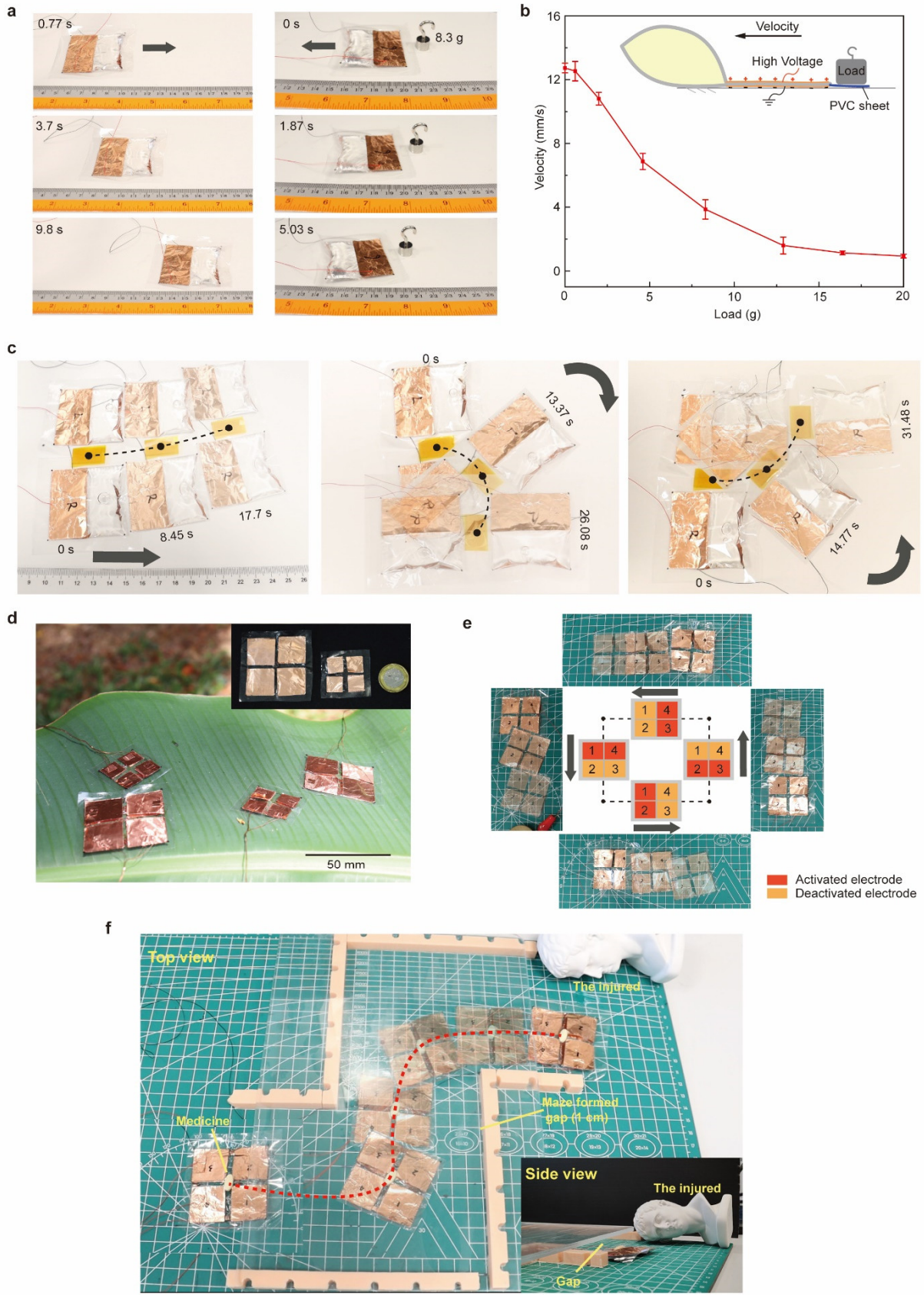

**Figure 5.** The single-DoF MCR and 2-DoF MCR. (a) The single-DoF MCR drags a load. (b) The crawling velocities with different loads. (c) The reconfigured 2-DoF MCR is composed of two MCRs. (d) The 4-electrode-MCR. (e) The four crawling directions of the 4-electrode-MCR. (f) The 4-electrode-MCR passes through a winding gap and delivers medicine to the injured.

**2.5. Underwater Locomotion**

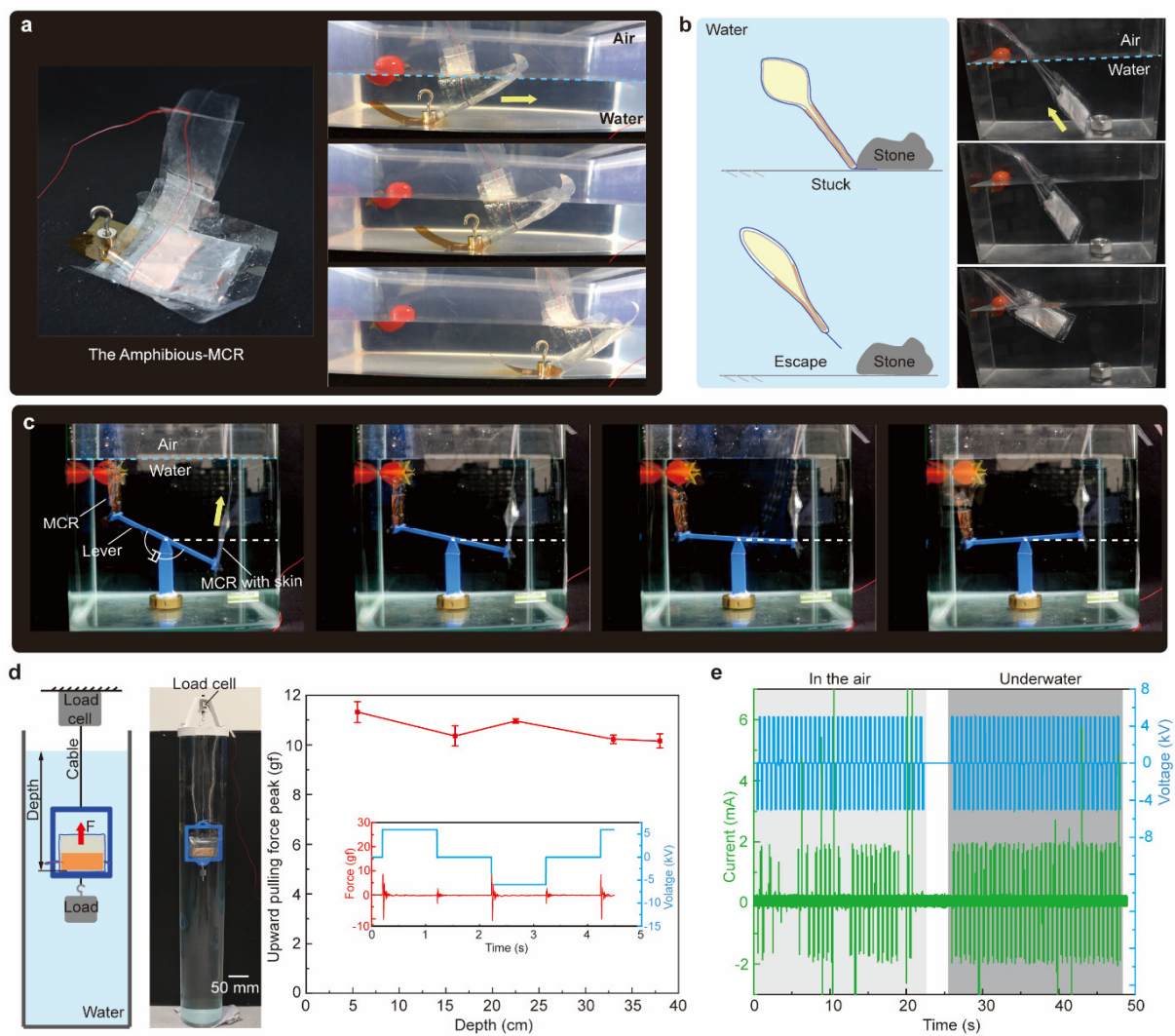

**Figure 6.** Underwater locomotion of the MCR. (a) The MCR with water-proof skin crawls underwater dragging a load. (b) The MCR escapes from being stuck. (c) The MCR rotates a lever. (d) The influence of water depth on EFKE. (e) The output current of HV supply.

With the waterproof skin, the MCR can crawl underwater (**Figure 6**a and Movie S2). With principle shown in Figure 1c, an 8.3 g weight was connected to the MCR tail allowing it to sink to the bottom (4 cm depth). The robot dragging the weight crawls step by step powered by EFKE. In addition, we showcased the ability to escape a larger load (63 g) with repeated generation of EFKE, when the MCR tail was pinned down (Figure 6b). Additionally, the underwater EFKE can be utilized to actuate other mechanisms under water. Here, we built an underwater lever with the pivot point in the middle allowing for free rotation. One end of the lever was connected to the water-proof MCR (right), and the other end was connected to another

MCR with water-proof skin (left) for balance. By powering the water-proof MCR, the generated upward EFKE repeatedly struck the bar and gradually rotated the lever (Figure 6c and Movie S9).

We further explored the influence of water depth on the underwater EFKE. The MCR with skin was fixed on a rigid frame that is connected to a load cell via a cable (Figure 6d). The rigid frame together with the MCR was immersed in the water connect to a 100g weight pulling the frame down to straighten the cable. The load cell can measure the actual force variation due to the underwater EFKE. We change the immersing depth from 5.5 to 38 cm and record the variation peak of the force which also implicitly indicates the EFKE. The results (Figure 6d) show that water depth cannot influence the EFKE. The whole volume of the MCR keeps constant owing to the liquid incompressibility. Assuming the pressure on the surface of the robot is equal everywhere, the robot hardly transfers energy to the water. Therefore, the water depth cannot influence the EFKE of the MCR with water-proof skin. Furthermore, we also compared the output current of the HV supply between the in-air and underwater MCRs. The water environment influences the operating current of the MCR (Figure 6e). The current peaks of underwater MCR were denser than those in the air likely due to the water increasing the capacitance.

**2.6. Robustness, Recovery and Durability of MCR**

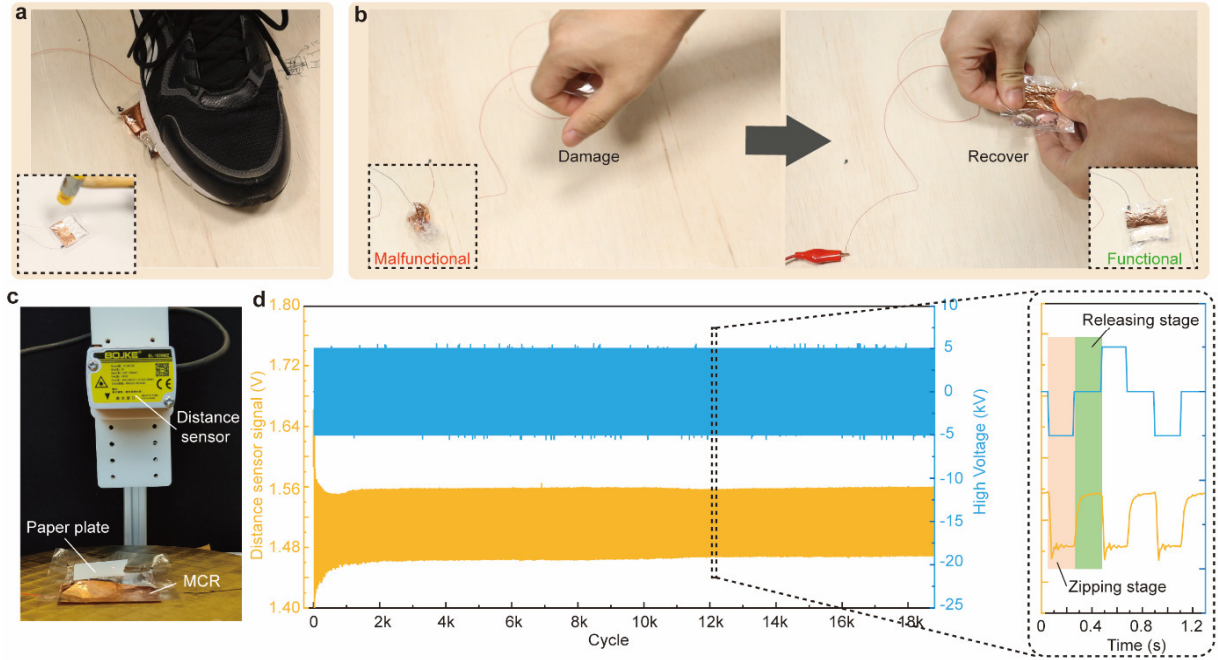

**Figure 7.** Robustness, recovery and durability test. (a) Step on the MCR with human feet and hammer impact on the MCR. (b) Recover the MCR after damage. (c) The platform for durability test. (d) The durability test results.

Benefiting from the fully-soft body, our MCR possesses high robustness. It can work normally despite being stepped on (**Figure 7**a and Movie S10) by a human. Even hammer impact cannot damage the structure of MCR due to its highly deformable components (liquid and films). In addition, our MCR has high recovery stemming from its compliant components (electrodes, the dielectric bladder and internal liquid). We intentionally knead the MCR trying to cause a malfunction. In spite of that, a simple stretch can recover its function (Figure 7b and Movie S11). The durability of MCR was also tested by the testing platform shown in Figure 7c. We adhered a piece of paper on top of the MCR and used a laser displacement sensor to measure the deformation of electrohydraulic actuator. A 5 kV AC HV (ZT/RT: 200/200 ms) powered the electrohydraulic actuator. Our MCR can run continuously for about 19k cycles without attenuation (more than 2 hours).

## 3. Conclusion

**Table 1.** Comparison with other soft miniature crawling robots

| Soft crawling robot | Actuation type | Softness | Average velocity (BL/s) | Body Length (mm) | Body weight (g) | Amphibious ability | Untethered |
|---|---|---|---|---|---|---|---|
| Kim et al.[36] | Electrohydraulic + Peristaltic mechanism | Full | 0.0011 | 125 | 8 | No | No |
| Wang et al.[42] | Electroadhesion | Part | 0.57 | 46 | 0.23 | No | No |
| Nan et al.[43] | IPMCs[a] | Part | 0.016/0.15[b] | 28 | 2.7 | High | Yes |
| Ji et al.[24] | DEA | Part | 0.3 | 40 | 1 | No | Yes |
| Huang et al.[44] | SMA | Part | 0.56 | 57 | 25 | No | Yes |
| Manfredi et al.[19] | Pneumatics | Full | 0.047 | 60 | 10 | High | No |
| This work | Electrohydraulic + EFKE | Full | 0.32 | 50 | 6.3 | Medium[c] | No |

a) Ionic polymer–metal composites actuator; b) 0.016 is for the crawling state and 0.15 is for the swimming state; c) The MCR equipped with water-proof skin;

In this work, we designed a fully-soft miniature crawling robot base on the directly-generated EFKE by electrohydraulic actuator. The MCR consisted of only an electrohydraulic actuator which was made from pre-coated BOPP membranes with lower heating temperature and reliable performance. After comprehensively optimizing the operating HV, the volume of dielectric liquid and the length of electrodes, our MCR achieved 16 mm/s crawling speed (0.32 body length per second). The optimized MCR weighs 6.3 g and has a 50×50×6 mm dimension, which is still scaleable. We also proposed two simple ways to reconfigure our MCR for 2-DoF locomotion which has immense potential in rescue and delivery operations. To extend its environmental adaptability, we designed a soft water-proof skin fitted onto the MCR which allows our MCR to locomote and actuate underwater. The fully-soft and deformable structure gave the robot favorable robustness and recovery that preserved the robot's functionality. Moreover, the MCR exhibited high durability verified from our experiments.

In contrast with other soft miniature crawling robots (**Table 1**), our MCR with full softness can achieve relatively high crawling velocity. The EFKE-driven mechanism considerably improves the crawling speed, almost 290 times that of the other MCR actuated by electrohydraulic actuators. Most other MCRs require rigid leg or body architectures, which limits the softness of the robots. Moreover, our soft MCR equipped with soft water-proof skin has a certain amphibious ability.

Future work involves further exploring the amphibious ability, such as seamlessly transitioning between water and land and improving the swimming ability by integrating soft paddles onto the MCR. New dielectric materials can also be explored with the aim of reducing the operating voltage. In addition, an untethered MCR can be explored using a compact and lightweight power on-board circuit.

## 4. Experimental Methods

*Circuit system*: The MCR is powered by a high voltage supply (Trek 10/40, Advanced Energy) via 0.3 mm diameter Teflon electric wires. The high voltage supply's ports are connected to a multifunction DAQ device (USB-6211, National Instruments). The DAQ device is programmed by LabVIEW 2018. We can control the ZT, RT and the amplitude of HV by regulating the parameters in the LabVIEW procedure.

*Velocity test*: To test the crawling velocity, we set the total operating time to 5 s, and measured the motion distance by a rule to calculate the average crawling velocity. The slow-motion videos were shot by a cellphone (Realme GT 2Pro, OPPO).

## Supporting Information

Supporting Information is available from the Wiley Online Library or from the author.

**Acknowledgements**

This work was funded by A*STAR Industry Alignment Fund – Pre-Positioning (A20H8A0241).

Received: ((will be filled in by the editorial staff))
Revised: ((will be filled in by the editorial staff))
Published online: ((will be filled in by the editorial staff))

**Comments for Movies**

Movie S1: The MCR is pushed by EFKE

Movie S2: The MCR with water-proof skin can crawl underwater

Movie S3: The MCR with water-proof skin can jump underwater

Movie S4: The amplitude of HV influences the stride of MCR

Movie S5: The MCR can crawl at a fast speed

Movie S6: The combined MCR can move and turn

Movie S7: The 4-electrode-MCR can crawl multi-directionally

Movie S8: The 4-electrode-MCR transports medication while passing through a winding gap to the injured

Movie S9: The MCR with water-proof skin can rotate the lever

Movie S10: The MCR can work normally despite being stepped on by a human, and hammer impact

Movie S11: The MCR has high recovery